\pdfoutput=1

\documentclass[11pt]{article}

\usepackage[final]{ACL2023}
\usepackage{times}
\usepackage{latexsym}

\usepackage[T1]{fontenc}

\usepackage[utf8]{inputenc}

\usepackage{microtype}

\usepackage{inconsolata}
\usepackage{url}
\usepackage{algorithm}
\usepackage{algorithmic}

\usepackage{amsmath}
\usepackage{amssymb}
\usepackage{tcolorbox}
\usepackage{multirow}
\usepackage{multicol}
\usepackage{booktabs}
\usepackage{tabularx}

\title{Neuro-Symbolic Query Compiler}

\author{%
Yuyao Zhang$^1$, Zhicheng Dou$^1$\thanks{Correpsonding author.}, Xiaoxi Li$^1$, Jiajie Jin$^1$\\
\textbf{Yongkang Wu$^2$, Zhonghua Li$^2$, Qi Ye$^2$} and \textbf{Ji-Rong Wen$^1$} \\
$^1$Renmin University of China\\
$^2$Huawei Poisson Lab\\
\texttt{\{2020201710, dou\}@ruc.edu.cn}
}

\newcommand{\ourmodel}{\textbf{QCompiler}}

\begin{document}
\thispagestyle{plain}
\maketitle

\begin{abstract}
    Precise recognition of search intent in Retrieval-Augmented Generation (RAG) systems remains a challenging goal, especially under resource constraints and for complex queries with nested structures and dependencies. This paper presents \ourmodel, a neuro-symbolic framework inspired by linguistic grammar rules and compiler design, to bridge this gap. It theoretically designs a minimal yet sufficient Backus-Naur Form (BNF) grammar $G[q]$ to formalize complex queries. Unlike previous methods, this grammar maintains completeness while minimizing redundancy. Based on this, QCompiler includes a Query Expression Translator, a Lexical Syntax Parser, and a Recursive Descent Processor to compile queries into Abstract Syntax Trees (ASTs) for execution. The atomicity of the sub-queries in the leaf nodes ensures more precise document retrieval and response generation, significantly improving the RAG system's ability to address complex queries.
\end{abstract}
\section{Introduction}
In the field of cognitive science, the human brain demonstrates a sophisticated synergy between two cognitive systems \cite{hitzler2022neuro}: On the one hand, through \textbf{neural-networks-based computation}, humans can rapidly process information from complex sensory inputs. On the other hand, with \textbf{symbolic-system-based logical reasoning}, humans can analyze abstract rules such as language, mathematics, and causality, performing calculations and inferences through symbols and rules. These two systems complement each other, enabling humans to flexibly handle a range of complex tasks from perception to reasoning, exhibiting a powerful generalization capability that cannot be achieved by a single mechanism alone.

\begin{figure}[!t]
    \centering
    \includegraphics[width=1\linewidth]{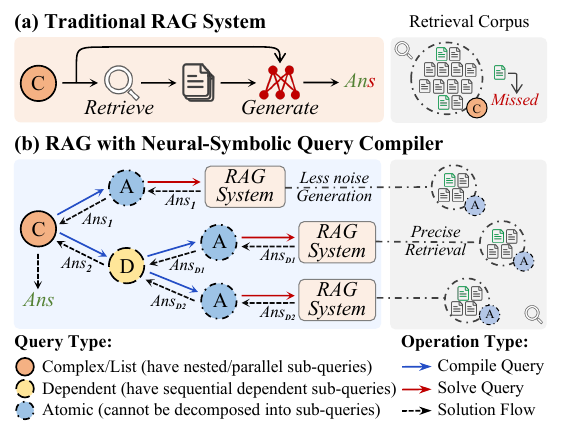}
    \caption{
    Illustration of how \ourmodel{} enhances the RAG system.
    }
    \label{fig:compare} 
\end{figure}

In the field of Artificial Intelligence, Artificial Neural Networks (ANNs) have shown strong fitting capabilities but struggle to extrapolate and generalize in a world of constantly updated knowledge~\cite{hasson2020direct}. Recently, \textbf{R}etrieval-\textbf{A}ugmented \textbf{G}eneration (RAG) technique addresses this limitation to some extent by introducing a retrieval process that allows ANNs to access external knowledge beyond their training data~\cite{lewis2020retrieval,zhao2024retrieval}. This enhancement improves the ability of neural networks to process information in open domains and real world scenarios. However, this improvement has a limited upper bound: As user queries become more complex or require reasoning, the chance of retrieving all relevant documents at once decreases significantly, leading to poor performance of RAG systems, as shown in Figure \ref{fig:compare}(a).

The handling of complex user queries for ANNs presents a significant challenge: These queries often contain \textbf{implicit intents}, \textbf{nested logical structures}, and \textbf{intricate dependencies}, making it difficult to arrive at answers in a single step. For example, the query “I want to find an introduction and reviews of J.K. Rowling’s most popular book and check if the local library has it” may require the system to use ANNs to extract key information, while relying on symbolic rules for task decomposition and reasoning—enabling multi-turn interaction with both databases and users. The system cannot execute other queries without first identifying which is the most popular book. This naturally raises the first question: \textbf{How can we effectively process complex queries to recognize search intent precisely?}

A straightforward idea is to leverage the powerful ANNs' capabilities to achieve this in an end-to-end manner~\cite{ma2023query,selfrag,rqrag}. Unlike the human brain, however, ANNs often struggle with tasks that require \textbf{explicit symbolic reasoning}. This is in part because they may not encounter sufficient data related to specific symbolic reasoning, leading to implicit reasoning modeling, making it difficult to handle complex queries that require explicit analysis and decomposition. This raises a second question: \textbf{How can we replicate the synergy of two cognitive systems in human brain, neural computation and symbolic reasoning, to effectively address complex queries in real-world scenarios?}

Many studies have investigated approaches for rewriting, disambiguating, and decomposing complex queries using symbolic and heuristic rules~\cite{min2019multi,khot2020text,rqrag}. We can regard complex queries as a Domain Specific Language (DSL) generated by a certain complete grammar $G$~\cite{wang2024grammar}, and such approaches can be viewed as selecting implicitly or explicitly a subset $G'$ of grammar $G$ (denoted as $G’ \subseteq G$) to generate queries. There has also been research on the use of LLMs to handle complex queries, with iterative and agentic RAG systems showing strong problem planing \& solving capabilities~\cite{verma2024plan,chen2024mindsearch}.

However, these methods face two main issues: (1) Although the processing of complex queries may cover all implied intentions, from the perspective of grammar-generated languages, this coverage may not be minimal~\cite{wolfson2020break}, resulting in unnecessary complexity and resources waste. (2)LLM-based iterative and agentic RAG systems rely heavily on their performance. This reliance often worsens the trade-off between performance and efficiency: achieving high performance typically requires redundant retrievals or frequent API calls for multi-round iterations, leading to higher computational costs and increased latency.

To address these questions, we propose \ourmodel, inspired by grammars in linguistics and compiler theory. We first summarize four query types and design a Backus-Naur Form (BNF) grammar $G[q] \subset G' \subseteq G$ that is minimally sufficient to formalize these queries. Then we train a small language model to translate natural-language queries into these BNF-based expressions, which can be parsed into an Abstract Syntax Tree (AST). The grammar G[q] constrains the search space during language model generation, while the parsed abstract syntax tree (AST) enables symbolic reasoning by recursively applying grammar rules at each node. This process efficiently handles nested structures and dependencies of complex queries, resembling a compiler that can translate high-level code into machine-executable instructions.

This framework can be seamlessly customized into existing RAG system for better query understanding because: (1) The parsed AST can capture the implicit search intents, nested structure, and dependencies of original complex query. (2) The sub-queries in leaf nodes are more precise because they represent well-defined atomic sub-queries, reducing ambiguity and targeting specific intent for accurate document retrieval and generation, as shown in Figure \ref{fig:compare}(b). (3) It allows developers in production environments to verify the correctness of each sub-query node and, when necessary, modify or intervene in the reasoning process. The results across four multi-hop benchmarks show QCompiler's remarkable understanding capability of complex queries.

Our contributions in this paper are as follows:

1. We theoretically propose a minimal yet sufficient Backus-Naur form (BNF) grammar $G[q]$ to formalize complex queries and provide a foundation for precise search intent recognition.

2. We propose \ourmodel{}, which synergize neural computation and symbolic reasoning to compile complex queries into Abstract Syntax Trees, capturing their nested structures and dependencies.

3. We experimentally demonstrate the accuracy of this tree structure representations in expressing complex queries, which improves efficiency and accuracy by retrieving fewer but more precise documents and providing more accurate responses.
\section{Related Work}

\subsection{Query Understanding}
Query understanding involves rewriting, disambiguating, decomposing, and expanding the original query methods~\cite{ma2023query,min2019multi,khot2020text,selfrag,rqrag}. Approaches to handling complex multi-hop queries focus mainly on the decomposition rule~\cite{min2019multi,wolfson2020break}.To enhance the ability of Question Answering, researchers focuses on constructing benchmarks for complex multi-hop questions, either manually or using symbolic rules, to evaluate existing systems~\cite{hotpotqa,2wiki,Musique}. The main solutions currently are mainly focused on the use of relevant documents and parametric knowledge of LLMs to process queries~\cite{selfrag,ircot,iteretgen,rqrag}. QCompiler integrates neural networks and reasoning based on symbolic rules. It naturally includes the aforementioned aspects of query understanding during the parsing and execution of complex queries.

\subsection{Neuro-Symbolic AI \& Grammar Rule}
Neuro-symbolic AI combines the computational power of neural networks with the precise rule-based reasoning of symbolic systems~\cite{hitzler2022neuro,LINC,ahmed2022semantic,dinu2024symbolicai}. This paradigm has shown significant potential in the treatment of complex and even long-tail tasks~\cite{visualprogramming,yao2022react,schick2024toolformer,shen2024hugginggpt}. Recent studies highlight the benefits of parameterizing symbolic rules, such as grammars, to enhance ANN's symbolic reasoning capabilities in specific domains~\cite{dinu2024symbolicai,wang2024grammar}. Drawing inspiration from lexical and syntactic analysis in context-free grammars and compiler design~\cite{wang2024grammar,aho1969translations,alfred2007compilers}, QCompiler introduces a parameterized grammar-based model to compile complex queries. This approach progressively translates queries into intermediate expressions and parses them into ASTs, achieving efficient parsing, reasoning, and inference.

\subsection{Retrieval-Augmented Generation System}
RAG systems integrate retrieval and generative models to access external knowledge during inference~\cite{lewis2020retrieval,zhao2024retrieval}, overcoming the static nature of parametric models. More advanced approaches incorporate modules for query rewriting~\cite{ma2023query}, retrieval necessity determination~\cite{tan2024small,jiang2023active}, document re-ranking, refinement~\cite{jiang2023longllmlingua,li2023compressing}, and others~\cite{jin2024flashrag}. Iterative and agentic RAG systems can also perform tree-structured~\cite{rqrag} or graph-structured~\cite{chen2024mindsearch,li2024benchmarking} reasoning during processing by leveraging mechanisms such as reflection~\cite{selfrag} and planning~\cite{ircot,iteretgen}. However, challenges such as redundancy, high computational costs, and reliance on the performance of the base model still persist. 
\begin{figure*}[htbp]
    \centering
    \includegraphics[width=.98\linewidth]{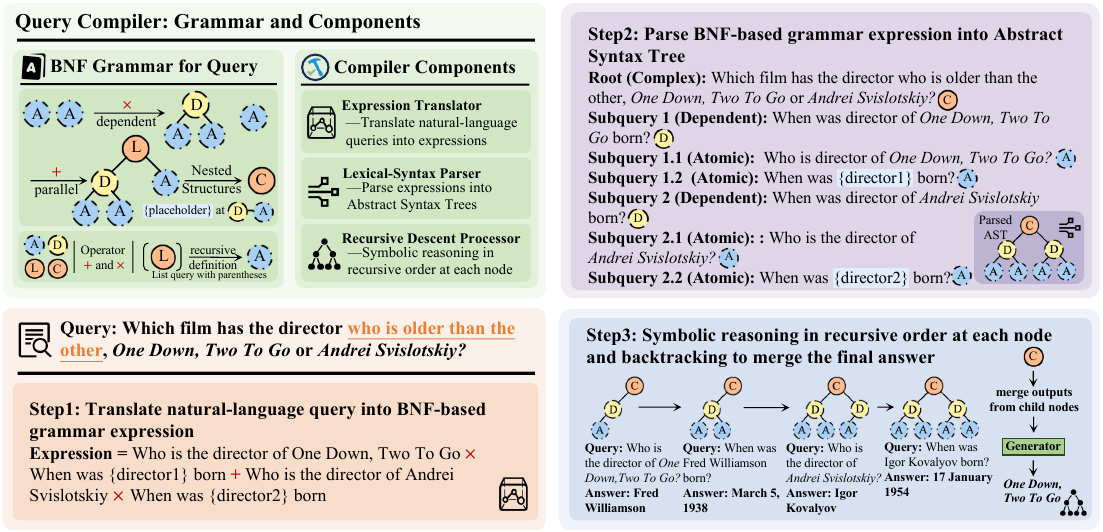}
    \caption{
    Illustration of \ourmodel, including the grammars, components, and an example of processing complex queries with QCompiler.
    }
    \label{fig:main} 
\end{figure*}
\section{Methodology}
In this work, we focus on the understanding and representation of complex queries. We give an overview of \ourmodel{} and how it works in Figure \ref{fig:main}, and details of grammar, components used to implement each step, training, and inference in the following sections.

\subsection{Mathematical Definition of Query Types}
\label{section 3.1}
To effectively process complex queries and capture user intentions, we conceptualize different queries as four basic types: \textbf{\textit{Atomic Query}}, \textbf{\textit{Dependent Query}}, \textbf{\textit{List Query}}, and \textbf{\textit{Complex Query}}. Let $Q$ be the set of all possible queries and:

\begin{tcolorbox}[
    colframe = blue!50!black!75!,       
    colback = gray!5!white,             
    coltitle = white,                   
    coltext = black,                    
    fonttitle = \bfseries,              
    title = DEFINITION 1 ,              
    boxrule = 1pt,                      
    arc = 1mm,                          
    width = \linewidth,                 
    left = 5pt,                         
    right = 5pt,                        
    top = 5pt,                          
    bottom = 5pt                        
]
\fontsize{9.5pt}{11.5pt}\selectfont
    \textbf{$<$\textit{Atomic Query}$>$} A query $q \in Q$ is called an \textbf{atomic query} if it cannot be decomposed into a combination of sub-queries. For $q, \forall q_1, \cdots q_n\in Q$, s.t. 
    $$
        q \neq q_1 \circ q_2 \circ \dots \circ q_n
    $$
    where $\circ$ denotes a query composition operation.
\end{tcolorbox}

In other words, an atomic query is an indivisible single-turn query that cannot be further decomposed. It requires no external context or dependencies for its execution, such as "\textit{Who is the director of The Titanic?}".

\begin{tcolorbox}[
    colframe = blue!50!black!75!,       
    colback = gray!5!white,             
    coltitle = white,                   
    coltext = black,                    
    fonttitle = \bfseries,              
    title = DEFINITION 2,               
    boxrule = 1pt,                      
    arc = 1mm,                          
    width = \linewidth,                 
    left = 5pt,                         
    right = 5pt,                        
    top = 5pt,                          
    bottom = 5pt                        
]
\fontsize{9.5pt}{11.5pt}\selectfont
    \textbf{$<$\textit{Dependent Query}$>$} A query $ q $ is called a \textbf{dependent query} if it can be decomposed into two parts. For $q, \exists \ q_1, q_2 \in Q$, s.t.
    $$
        q = q_1 \circ q_2, \text{and } q_2 = f(q_1)
    $$
    where $ f $ is a function that maps the result of $q_1$ to $q_2$.
\end{tcolorbox}
A dependent query consists of two parts, with the latter part's execution reliant on the results of the former part and cannot be executed in parallel, such as "\textit{When was the director of The Titanic born?}" containing two queries with dependencies: "\textit{Who is the director of The Titanic?}" and "\textit{When was James Cameron born?}".

\begin{tcolorbox}[
    colframe = blue!50!black!75!,       
    colback = gray!5!white,             
    coltitle = white,                   
    coltext = black,                    
    fonttitle = \bfseries,              
    title = DEFINITION 3,               
    boxrule = 1pt,                      
    arc = 1mm,                          
    width = \linewidth,                 
    left = 5pt,                         
    right = 5pt,                        
    top = 5pt,                          
    bottom = 5pt                        
]
\fontsize{9.5pt}{11.5pt}\selectfont

    \textbf{$<$\textit{List Query}$>$} A query $ q $ is called a \textbf{list query} if it can be decomposed into multiple indenpendent sub-queries. For $q, \exists \ q_1, \cdots q_n\in Q$, s.t.
    $$
        q = q_1 \circ q_2 \circ \dots \circ q_n
    $$
    and for $\forall i \neq j$, $q_i \neq f(q_j)$.

\end{tcolorbox}

Thus, a list query consists of multiple parts that are independent and can be executed in parallel to accelerate the inference of a whole system, such as "\textit{Who is older, James Cameron or Steven Allan Spielberg?}" containing two queries without dependencies: "\textit{When was James Cameron born}" and "\textit{When was Steven Allan Spielberg born}".

\begin{tcolorbox}[
    colframe = blue!50!black!75!,       
    colback = gray!5!white,             
    coltitle = white,                   
    coltext = black,                    
    fonttitle = \bfseries,              
    title = DEFINITION 4,               
    boxrule = 1pt,                      
    arc = 1mm,                          
    width = \linewidth,                 
    left = 5pt,                         
    right = 5pt,                        
    top = 5pt,                          
    bottom = 5pt                        
]
\fontsize{9.5pt}{11.5pt}\selectfont
    \textbf{$<$\textit{Complex Query}$>$} A query is called a \textbf{complex query} if it can be decomposed into multiple sub-queries, which may include atomic, dependent, and list queries. For $ q, \exists \ q_1, q_2, \dots, q_n \in Q $, s.t.
    $$
        q = q_1 \circ q_2 \circ \dots \circ q_n
    $$
    and for at least one pair $(i, j)$, st. $q_i = f(q_j)$.
\end{tcolorbox}

Thus, a complex query involves combining multiple types of queries with nested logical structure and intricate dependencies, forming a structured query that cannot be classified solely as an atomic query, dependent query, or list query, such as \textit{Who is older, the director of The Titanic or Steven Allan Spielberg?} contains a dependent query and an atomic query without dependencies.

Under this definition, all queries in real-world scenarios can be described and formalized.

\subsection{Backus-Naur Form (BNF) Grammar}
\label{section 3.2}
With the query definition mentioned above, in this section we present the specialized Backus-Naur Form (BNF) grammar for complex queries.

BNF is a context-free grammar widely used to precisely describe syntax rules in programming languages, protocols, and Domain-Specific Languages (DSLs)~\cite{aho1969translations,alfred2007compilers,wang2024grammar}. Following these works, a BNF grammar $G$ is typically defined as a $4$-tuple
    $G = (N, T, P, S)$
, where:

(1) $N$ is a set of non-terminal symbols that represent intermediate structures in grammar.

(2) $T$ is a set of terminal symbols, representing concrete characters or tokens in the language, with $N \cap T = \emptyset$.

(3) $S \in N$ is the start symbol.

(4) $P$ is a set of production rules, each of the form $A \to \alpha$, where $A \in N$ and $\alpha \in (N \cup T)^*$.
Each production rule $A \to \alpha$ is often written as:
$
   \langle A \rangle ::= \alpha_1 \;|\; \alpha_2 \;|\; \dots \;|\; \alpha_k, 
$
where $\alpha_i \in (N \cup T)^*$ represents possible expansions of $A$. 

Then we define the BNF grammar for complex queries according to this paradigm as follows:

\textbf{Non-terminal symbols $\boldsymbol{N}$.} Non-terminals denote intermediate structures of grammar, abbreviated as $<$\textit{\textbf{Atomic}}$>$, $<$\textit{\textbf{List}}$>$, $<$\textit{\textbf{Dependent}}$>$, $<$\textit{\textbf{Complex}}$>$. 

\textbf{Terminal symbols $\boldsymbol{T}$.}  It is divided into two subsets: the atomic query set $Q_{\mathrm{atomic}} \subset Q$ and the operator set $O$. 
Each $q \in Q_{\mathrm{atomic}}$ represents an atomic query string, such as "\textit{Who is the director of The Titanic?}".
The operator set $O$ contains `$+$' and `$\times$'. `$+$' connects two independent queries, allowing them to be answered in parallel; $\times$ connects two queries in a dependent relationship, where the latter query relies on the result of the former query. 

\textbf{Start Symbol $\boldsymbol{S}$.} For every query $q \in Q$, the start symbol corresponds to a \textbf{$<$\textit{Complex}$>$}, which serves as the root of the grammar.

\textbf{Production Rules $\boldsymbol{P}$.} Together with the definition in Section \ref{section 3.1}, we further refer to BNF to define the production rules for complex queries.
\begin{tcolorbox}[
    colframe = red!50!black,       
    colback = gray!5!white,             
    coltitle = white,                   
    coltext = black,                    
    fonttitle = \bfseries,              
    title = PRODUCTION RULES OF QUERY,  
    boxrule = 1pt,                      
    arc = 1mm,                          
    width = \linewidth,                 
    left = 5pt,                         
    right = 5pt,                        
    top = 5pt,                          
    bottom = 5pt,                        
]
\fontsize{8.6pt}{13pt}\selectfont
    $<$\textit{\textbf{Atomic}}$>$$::= q \in Q_{atomic}$ $|$ $($$<$\textit{\textbf{List}}$>$$)$ \\
    $<$\textit{\textbf{Dependent}}$>$$::=$$<$\textit{\textbf{Atomic}}$>$$|$$<$\textit{\textbf{Dependent}}$>$$\times$$<$\textit{\textbf{Atomic}}$>$ \\
    $<$\textit{\textbf{List}}$>$$::=$$<$\textit{\textbf{Dependent}}$>$$|$$<$\textit{\textbf{List}}$>$$+$$<$\textit{\textbf{Dependent}}$>$ \\
    $<$\textit{\textbf{Complex}}$>$$::=$ $<$\textit{\textbf{List}}$>$
\end{tcolorbox}

In production rules, the operator `$\times$' is assigned a higher precedence than the operator `$+$' to ensure that the parsing process is deterministic and unambiguous. Furthermore, we use parentheses for grouping and priority control, and the \textbf{ expressions enclosed in parentheses can also be considered as a production rule for atomic queries}. This recursive definition is similar to those found in many programming languages and general-purpose grammars, allowing nested queries to be formalized naturally without complicating the grammar or introducing additional non-terminal variables.

We provide a proof of the completeness and minimality of grammar $G[q]$ in Appendices \ref{appendix:completeness} and \ref{appendix:minimality}.

\subsection{Key Components of QCompiler}
\label{sec:component}
With the queries and their BNF grammar defined, the next step is to process complex queries. Drawing inspiration from compiler design, we develop a compiler instance that integrates three key components: a Query Expression Translator, a Lexical-Syntax Parser, and a Recursive Descent Processor. 

\textbf{Query Expression Translator.} This component uses a language model to translate natural language queries into BNF-based expressions. It ensures that the user's search intents should be precisely captured and represented. To achieve this, we train a language model to specifically translate natural language queries into BNF-based expressions, as illustrated in Figure~\ref{fig:main} Step1.

\textbf{Lexical-Syntax Parser.} This component performs symbolic reasoning on query expressions, using tokens from lexical analysis to construct an Abstract Syntax Tree (AST) based on the BNF grammar. The AST represents the structure of the query, capturing nested structure and dependency, and serves as a bridge between the representation of high-level query and the downstream reasoning processes, as illustrated in Figure~\ref{fig:main} Step2.

\textbf{Recursive Descent Processor.} This component interprets AST recursively, executing the sub-queries by resolving their dependencies and performing placeholder replacements. It manages the data flow between different query nodes, handling execution of sub-queries in the AST, as illustrated in Figure~\ref{fig:main} Step3.

\subsection{Training of Query Expression Translator}
\label{section 3.4}
To enable the language model to understand the grammar and respond in the desired format, we optimize it using the following objective function:
\begin{equation}
    \mathcal{L} = - \sum_{\langle q_i, e_i \rangle \in \mathcal{D}_T} \log P(e_i \mid q_i, G[q]),
\end{equation}
where $\mathcal{D}_T$ represents the training dataset consisting of query-expression pairs $\langle q_i, e_i \rangle$, $q_i$ is the input query, $e_i$ is the corresponding expression following the grammar $G[q]$ and $G[q]$ denotes the BNF grammar instruction. The construction of training data is detailed in Appendix~\ref{appendix:train}. The trained model can translate original query into BNF-expression. 

\begin{table*}[htbp]
\renewcommand\arraystretch{1.0}
\centering
\scalebox{.9}{
\begin{tabular}{l|ccc|ccc|ccc|ccc}
\toprule
\multirow{2}{*}{\textbf{Methods}} 
& \multicolumn{3}{c|}{\textbf{2Wiki}} 
& \multicolumn{3}{c|}{\textbf{HotpotQA}} 
& \multicolumn{3}{c|}{\textbf{Musique}} 
& \multicolumn{3}{c}{\textbf{Bamboogle}} \\ 
\cmidrule(lr){2-4} \cmidrule(lr){5-7} \cmidrule(lr){8-10} \cmidrule(lr){11-13} & EM & Acc & F1 & EM & Acc & F1 & EM & Acc & F1 & EM & Acc & F1 \\ 
\midrule
\multicolumn{13}{l}{\textit{\textbf{Direct Generation}}} \\ 
Qwen2.5-7B-Instruct & $26.9$ & $28.5$ & $32.2$ & $17.2$ & $20.2$ & $26.7$ & $5.9$ & $9.1$ & $13.6$ & $8.8$ & $14.4$ & $16.9$ \\ 
\midrule
\multicolumn{13}{l}{\textit{\textbf{Sequential RAG System}}} \\ 
Naive RAG (TopK = 5) & $25.8$ & $29.5$ & $32.7$ & $33.8$ & $39.1$ & $44.9$ & $11.5$ & $15.9$ & $20.8$ & $18.4$ & $21.6$ & $28.3$ \\ 
Naive RAG (TopK = 10) & $28.2$ & $31.7$ & $34.8$ & $35.3$ & $39.2$ & $46.1$ & $12.8$ & $18.1$ & $20.0$ & $13.6$ & $15.2$ & $22.1$ \\
\midrule
\multicolumn{13}{l}{\textit{\textbf{Iterative RAG System}}} \\ 
Self-RAG & $11.3$ & $32.0$ & $23.5$ & $18.3$ & $34.0$ & $30.5$ & $6.3$ & $12.3$ & $15.1$ & $3.2$ & $16.8$ & $13.1$ \\ 
IR-CoT & $24.3$ & $\underline{37.3}$ & $34.4$ & $33.6$ & $\underline{42.6}$ & $45.6$ & $12.4$ & $19.7$ & $21.0$ & $21.6$ & $\underline{26.4}$ & $34.2$ \\ 
Iter-Retgen & $\underline{29.8}$ & $34.2$ & $36.2$ & $\underline{37.0}$ & $41.7$ & $\underline{49.1}$ & $15.3$ & $19.0$ & $22.9$ & $23.2$ & $24.0$ & $31.8$ \\ 
\midrule
\multicolumn{13}{l}{\textit{\textbf{Query Understanding}}} \\ 
RQ-RAG & $26.8$ & $35.8$ & $\underline{36.8}$ & $28.9$ & $33.5$ & $36.3$ & $\underline{19.8}$ & $\underline{24.8}$ & $\underline{27.8}$ & $\underline{24.8}$ & $\underline{26.4}$ & $\underline{34.7}$ \\ 
\rowcolor[RGB]{236,244,252}\ourmodel~(ours) & $\textbf{44.5}$ & $\textbf{53.5}$ & $\textbf{51.9}$ & $\textbf{38.5}$ & $\textbf{46.1}$ & $\textbf{50.2}$ & $\textbf{25.0}$ & $\textbf{35.4}$ & $\textbf{34.8}$ & $\textbf{37.1}$ & $\textbf{41.9}$ & $\textbf{49.2}$ \\ 
\bottomrule
\end{tabular}
}
\caption{Comparison of different methods. The best results are in bold. The base generator model is Qwen-2.5-7B-Instruct, and \ourmodel~is a fine-tuned Llama3.2-3B-Instruct model.}
\label{table:main_result}
\end{table*}

\subsection{Validation and Inference}
\label{section 3.5}
The trained Query Expression Translator may still generate invalid expressions, leading to the construction of invalid ASTs. These problems can be categorized into two types: (1) \textbf{Erroneous Dependency}: Placeholder content appears in query nodes without the corresponding dependencies. (2) \textbf{Missing Dependency}: Query nodes with dependencies lack the necessary placeholders to extract the required information. 

To address these issues, we designed a recursive validation algorithm based on Depth-First Search (DFS) to check the legality of ASTs, which is illustrated in Appendix~\ref{algo:valid}. During inference, we sample the outputs at various temperature settings and then select a valid AST for subsequent processing. 

\subsection{Why Can \ourmodel{} Improve Existing RAG Systems?}
QCompiler can improve RAG systems in multiple ways: (1) Unlike existing end-to-end approaches, QCompiler is a lightweight framework that focuses on generating \textbf{structured intermediate representations} for complex queries by compiling them into ASTs to capture their implicit intentions, nested structures, and intricate dependencies. This process inherently handles the rewriting, disambiguation, decomposition, and expansion of complex queries. (2) The atomicity of the sub-queries in the leaf nodes ensures precise document retrieval and answer generation, significantly improving the RAG system’s ability to address complex queries. (3) In practical deployment scenarios, developers can even design extensive post-processing logic to refine the AST compiled by QCompiler. These features make QCompiler highly adaptable to integration with existing RAG systems.
\section{Experiment}
\begin{table*}[htbp]
\renewcommand\arraystretch{1.0}
\centering
\scalebox{.9}{
\begin{tabular}{l|ccc|ccc|ccc|ccc}
\toprule
\multirow{2}{*}{\textbf{Methods}} 
& \multicolumn{3}{c|}{\textbf{2Wiki}} 
& \multicolumn{3}{c|}{\textbf{HotpotQA}} 
& \multicolumn{3}{c|}{\textbf{Musique}} 
& \multicolumn{3}{c}{\textbf{Bamboogle}} \\ 
\cmidrule(lr){2-4} \cmidrule(lr){5-7} \cmidrule(lr){8-10} \cmidrule(lr){11-13} & EM & Acc & F1 & EM & Acc & F1 & EM & Acc & F1 & EM & Acc & F1 \\ 
\midrule
\multicolumn{13}{l}{\textit{\textbf{Llama3.x series}}} \\ 
Llama3.2-1B-Instruct & $\textbf{45.1}$ & $\textbf{54.1}$ & $\textbf{52.5}$ & $36.9$ & $45.1$ & $49.0$ & $\textbf{25.1}$ & $35.3$ & $\textbf{35.2}$ & $34.7$ & $38.9$ & $46.7$ \\ 
Llama3.2-3B-Instruct & $44.5$ & $53.5$ & $51.9$ & $\textbf{38.5}$ & $\textbf{46.1}$ & $\textbf{50.2}$ & $25.0$ & $35.4$ & $34.8$ & $\textbf{37.1}$ & $41.9$ & $\textbf{49.2}$ \\ 
Llama3.1-8B-Instruct & $\underline{44.6}$ & $\underline{53.7}$ & $\underline{52.0}$ & $\underline{38.0}$ & $\underline{45.6}$ & $\underline{49.9}$ & $25.1$ & $\textbf{36.0}$ & $35.1$ & $36.6$ & $42.1$ & $49.2$ \\
\midrule
\multicolumn{13}{l}{\textit{\textbf{Qwen2.5 series}}} \\ 
Qwen2.5-0.5B-Instruct & $44.1$ & $53.0$ & $51.5$ & $37.2$ & $44.4$ & $48.8$ & $\underline{25.0}$ & $34.9$ & $34.8$ & $35.7$ & $41.1$ & $48.6$ \\ 
Qwen2.5-1.5B-Instruct & $44.3$ & $53.3$ & $51.8$ & $36.9$ & $44.4$ & $49.0$ & $24.8$ & $35.5$ & $35.0$ & $\underline{36.8}$ & $41.4$ & $48.7$ \\ 
Qwen2.5-3B-Instruct & $44.3$ & $53.5$ & $51.8$ & $37.2$ & $44.2$ & $48.8$ & $24.9$ & $\underline{35.6}$ & $\underline{35.1}$ & $35.0$ & $\underline{42.2}$ & $48.9$ \\ 
Qwen2.5-7B-Instruct & $43.9$ & $53.5$ & $51.8$ & $36.9$ & $44.5$ & $49.2$ & $24.8$ & $35.4$ & $35.2$ & $35.7$ & $\textbf{42.9}$ & $\underline{49.1}$ \\ 
\bottomrule
\end{tabular}
}
\caption{The performance of \ourmodel{} fine-tuned on different base models.}
\label{table:scaling}
\end{table*}

\begin{figure*}[!t]
    \centering
    \includegraphics[width=1\linewidth]{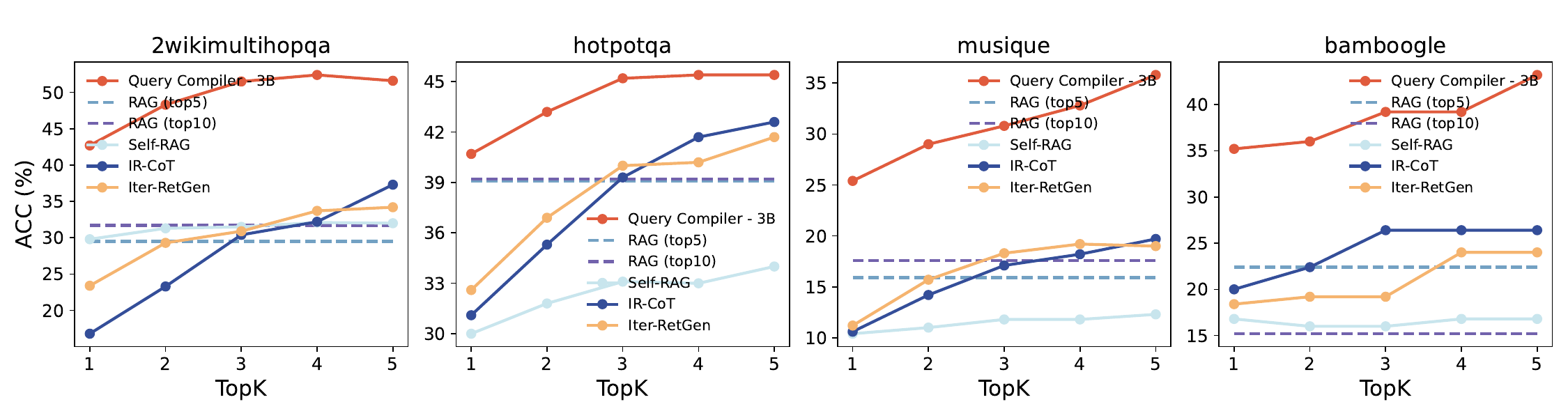}
    \caption{
    Illustration of the atomicity of sub-queries in leaf nodes, comparing with traditional RAG systems and Iterative RAG systems, \ourmodel{} has few documents retrieval and more accurate response.
    }
    \label{fig:topk} 
\end{figure*}

In this section, we validate the effectiveness of our method through a series of experiments.
\subsection{Datasets and Metrics}
We select four multi-hop benchmarks to validate the effectiveness of our approach in understanding complex queries. These datasets include \textbf{2WikiMultihopQA}~\cite{2wiki}, \textbf{HotpotQA}~\cite{hotpotqa}, \textbf{Musique}~\cite{Musique}, and \textbf{Bamboogle}~\cite{Bamboogle}. We use 2WikiMultihopQA, HotpotQA, and Musique's training set to construct QCompiler's fine-tuning datasets. We provide statistical descriptions of these benchmarks in Appendix~\ref{appendix:stat} and the training details of QCompiler in Appendix~\ref{appendix:train}. 

For evaluation, we use Extract Match (EM), Accuracy (Acc) and the F1 score to evaluate the results of the system responses.

\subsection{Baselines}
We mainly select the following four categories of baseline models:

\textbf{Direct Generation of LLM.} To better reflect the effectiveness of the retrieval augmentation process and the reasoning capabilities of complex RAG systems, we first compared the results of allowing the response model to directly generate answers.

\textbf{Sequential RAG System.} (1)~\textbf{Naive RAG}: Directly uses original query for retrieval and generation. Here we evaluate the generation with Top-5 and Top-10 retrieved documents separately for comparison.

\textbf{Iterative RAG System.} (1)~\textbf{Self-RAG}~\cite{selfrag}: Trains a language model to dynamically determine when to retrieve external information and critiques its outputs using specialized reflection tokens. (2)~\textbf{IR-CoT}~\cite{ircot}: Alternates between retrieval steps and Chain-of-Thought reasoning, allowing each step to inform and refine the other. (3)~\textbf{Iter-RetGen}~\cite{iteretgen}: Synergizes retrieval and generation iteratively, where model output guides subsequent retrievals, and retrieved information enhances future generations.

\textbf{Query Understanding.} (1)~\textbf{RQ-RAG}~\cite{rqrag}: Train a model to learn to rewrite, disambiguate, and decompose complex queries to retrieve documents and to generate the final response in an end-to-end manner.

We provide details on experiment implementations in the Appendix \ref{appendix:implementation}.

\subsection{Main Results}
Table \ref{table:main_result} presents the main results of various methods, including ours. ASTs compiled by QCompiler greatly improve the response model's capability without additional training for the model or retriever, achieving best performance on four benchmarks. Improvement is especially notable in challenging benchmarks like 2WikiMultihopQA and Musique.

\subsection{Analysis of Results}
The observed performance improvements can be attributed to the following factors:

\textbf{(1) Retrieval\&Generation Limitations of Original Queries.} Retrieval based on the original query often misses key information within the top-k documents due to insufficient context. Even larger k may fail to retrieve all relevant documents while introducing significant noise into the generation process, inherently limiting the performance of traditional RAG systems. 

\textbf{(2) Limited Iterative Query Planning.} For iterative RAG systems with constrained base model size or performance, the ability to plan new queries is weak, often requiring more iterations to complete a response. The sub-queries generated during these iterations can be viewed as derivations from a larger grammar $G'$ (rather than the minimal grammar $G[q]$) formalizing the original query, introducing redundancy for both the retrieved documents and the iterations, and further limiting the performance of iterative RAG systems. 

\textbf{(3) Improved Query Understanding with \ourmodel.} QCompiler is fine-tuned on a large dataset of grammar-based compiled expressions. It demonstrates a superior understanding of complex queries. It resembles generating a complete plan for the complex query in one step and performs reasoning based on predefined symbolic rules, which can provide an accurate answer to the query.

\begin{figure*}[!t]
    \centering
    \includegraphics[width=1\linewidth]{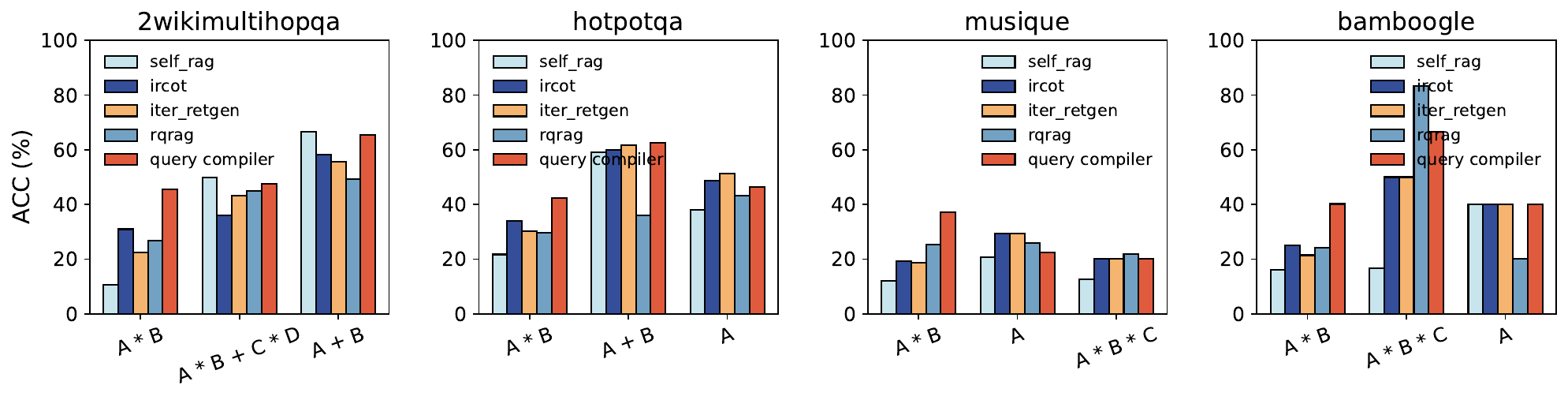}
    \caption{The number of correctly answered queries for each type compiled by \ourmodel{} across different methods.}
    \label{fig:cates} 
\end{figure*}
\section{Analysis}

In this section, we focus on analyzing how \ourmodel{} can improve the performance of the RAG system.

\subsection{Scaling law in \ourmodel}
We train QCompiler using base models of varying sizes with the same training data and experimental settings, and evaluate their performance in Table \ref{table:scaling}. The results reveal that performance across query compilers of different sizes is nearly identical. This suggests that the grammar-based generation task is relatively straightforward to learn, meaning that larger base models (e.g., Llama3.1-8B-Instruct and Qwen-2.5-7B-Instruct) do not necessarily outperform smaller ones (e.g., Llama3.2-3B-Instruct). This points to a limitation in current benchmarks for multi-hop queries, which may lack sufficient complexity and diversity, enabling smaller, distilled models to perform comparably well on existing benchmarks. 

\subsection{Analysis of Leaf Node's Atomicity}
An essential premise of QCompiler is the atomicity of sub-queries in leaf nodes. As illustrated in Figure \ref{fig:compare} and Section \ref{section 3.1},  a query represented by a leaf node in the AST should be an indivisible single-turn query that cannot be further decomposed into smaller components. For knowledge-intensive tasks, such single-turn queries must be precise enough to answer questions by retrieving as few documents as possible from the corpus. We compared the performance of different methods when retrieving different numbers of Top-K documents per query node, as shown in Figure \ref{fig:topk}. The results show that for QCompiler, retrieving only a small number of documents per query node (\textbf{even just one document}) is sufficient to achieve strong performance across different benchmarks. 

\subsection{Analysis of Query Types}
We use QCompiler to compile queries into their respective expression types. In Figure \ref{fig:cates}, we present data on the three most common query types, documenting the percentage of correct responses achieved for each type. The findings indicate: (1) QCompiler provides moderate improvement for single-hop questions since it applies a single cycle of refinement and processing in a recursive-descent manner, unlike iterative RAG systems that repeatedly refine these queries. (2) For list queries structured as $A + B$, QCompiler also provides a moderate improvement, suggesting that these queries are not difficult and can be managed by iterative RAG systems. (3) QCompiler excels with dependent queries, especially those formed as $A \times B$. This highlights the limitations of current iterative RAG systems: in multi-hop questions, the critical challenge lies in pinpointing the initial query and its answer correctly, a key factor that limits the system's effectiveness.
\section{Conclusion}
In this paper, we present \ourmodel{}, a neuro-symbolic framework inspired by linguistic grammar rules and compiler design. We first give a minimal yet sufficient grammar $G[q]$ to formalize and generate complex queries. Then QCompiler can efficiently compile each complex query into an Abstract Syntax Tree that contains these types and captures its nested structures and complex dependencies, demonstrating a strong ability to understand and analyze complex queries. The atomicity of the sub-queries in the leaf nodes ensures precise document retrieval and response generation. This lightweight framework enables seamless integration with existing RAG systems, highlighting its broad applicability and flexibility in practical deployment scenarios.

\section{Limitation}
Due to the limitations of existing multi-hop datasets, we lack more complex scenarios to train and validate the performance of the grammar-based QCompiler. For example, a key issue is the absence of benchmarks that feature complex queries that use parentheses to control the execution order, which may limit the generalization of the trained model. Additionally, in this paper, we focus solely on supervised fine-tuning to train QCompiler. Future improvement strategies include, but are not limited to, constructing more diverse and complex benchmarks for training and evaluation, and employing reinforcement learning with step-level reward models to generate more optimal expressions.

\bibliography{acl2023}
\bibliographystyle{acl_natbib}

\newpage
\section*{Appendix}
\appendix
\section{Proof of BNF Grammar's Completeness for Complex Queries}
\label{appendix:completeness}
Our goal is to prove that any complex query can be generated using the production rules of the given grammar.

\subsection{Lemmas}
\begin{tcolorbox}[
    colframe = blue!50!black!75!,       
    colback = gray!5!white,             
    coltitle = white,                   
    coltext = black,                    
    fonttitle = \bfseries,              
    title = LEMMAS,              
    boxrule = 1pt,                      
    arc = 1mm,                          
    width = \linewidth,                 
    left = 5pt,                         
    right = 5pt,                        
    top = 5pt,                          
    bottom = 5pt                        
]
\fontsize{8.5pt}{11.5pt}\selectfont
\textbf{Lemma1}: All \textbf{$<$\textit{Atomic Query}$>$} are \textbf{$<$\textit{Dependent Query}$>$}. 

\noindent\textbf{Lemma2}: All \textbf{$<$\textit{Dependent Query}$>$} are \textbf{$<$\textit{List Query}$>$}.
\end{tcolorbox}

According to the productions rules, these can be easily derived.
\subsection{Inductive Proof}

\begin{tcolorbox}[
    colframe = blue!50!black!75!,       
    colback = gray!5!white,             
    coltitle = white,                   
    coltext = black,                    
    fonttitle = \bfseries,              
    title = COMPLETENESS OF GRAMMAR,              
    boxrule = 1pt,                      
    arc = 1mm,                          
    width = \linewidth,                 
    left = 5pt,                         
    right = 5pt,                        
    top = 5pt,                          
    bottom = 5pt                        
]
\fontsize{9.5pt}{11.5pt}\selectfont
\textbf{Theorem}: Any valid query string constructed using query terms and operators `$+$` and `$*$` can be generated by the given grammar.
\end{tcolorbox}

\noindent\textbf{Proof}

\textbf{Base Case}:
Any atomic query $q \in Q_{atomic}$ is an \textbf{$<$\textit{Atomic Query}$>$} and also a \textbf{$<$\textit{Dependent Query}$>$}. Therefore, it is a \textbf{$<$\textit{List Query}$>$} and consequently a \textbf{$<$\textit{Complex Query}$>$}.

\noindent\textbf{Inductive Hypothesis}
Assume that the prodution rule can generate all valid query strings of length less than $n$. Here, the definition of length $n$ refers to a query containing $n$ atomic query strings.

\noindent\textbf{Inductive Step}
For a query string of length $n$, there are the following cases:

\textbf{Case 1}: The query is a dependent query in the form: `\textbf{$<$\textit{Dependent Query}$>$}$\times$\textbf{$<$\textit{Atomic Query}$>$}'.
By the inductive hypothesis, \textbf{$<$\textit{Dependent Query}$>$} and \textbf{$<$\textit{Atomic Query}$>$} can be generated by the production rule (because length $< n$). Therefore, this query can be generated by the production rule.

\textbf{Case 2}: The query string is a list query, in the form: `\textbf{$<$\textit{List Query}$>$}+\textbf{$<$\textit{Dependent Query}$>$}'.
By the inductive hypothesis, both \textbf{$<$\textit{List Query}$>$} and \textbf{$<$\textit{Dependent Query}$>$} can be generated by the production rule. Therefore, this query can be generated by production rule.

\textbf{Case 3}: The query string is a parenthesized list query in the form: `(\textbf{$<$\textit{List query}$>$})'. By \textbf{Case 1}, \textbf{$<$\textit{List Query}$>$} can be generated by the syntax. So the query can be generated by syntax.

\noindent\textbf{Conclusion}
By mathematical induction, query strings of length less than $n$, along with the inductive step for strings of length $n$, can be generated by the grammar. Therefore, the grammar is complete.

\section{Proof of Minimality of the Grammar}
\label{appendix:minimality}
To prove that the given grammar is minimal, we must show that: (1) The production rule generates the target query language. (2) No symbols or rules are redundant in the production rules. (3) Simpler production rules cannot generate the same language.

\subsection{Generation Completeness}
The completeness of the grammar has previously been proven, confirming that the grammar can generate all valid queries.

\subsection{Elimination of Redundant Symbols}
The non-terminal symbols in the grammars are:
\begin{itemize}
    \item \textbf{$<$\textit{Atomic Query}$>$}: Defines the basic single-turn query, indispensable.
    \item \textbf{$<$\textit{Dependent Query}$>$}: Introduces the `$\times$' operator for sub-queries with dependencies, which cannot be replaced by other terminal and non-terminal symbols.
    \item \textbf{$<$\textit{List Query}$>$}: Introduces the `$+$' operator for sub-queries without dependencies.
    \item \textbf{$<$\textit{Complex Query}$>$}: Serves as the starting symbol for the entire production rule and cannot be omitted.
\end{itemize}
Therefore, all symbols serve a distinct purpose and cannot be removed without reducing the generative power of the syntax.

\subsection{Elimination of Redundant Rules}
The production rules are as follows:
\begin{tcolorbox}[
    colframe = red!50!black,       
    colback = gray!5!white,             
    coltitle = white,                   
    coltext = black,                    
    fonttitle = \bfseries,              
    title = PRODUCTION RULES OF QUERY,  
    boxrule = 1pt,                      
    arc = 1mm,                          
    width = \linewidth,                 
    left = 5pt,                         
    right = 5pt,                        
    top = 5pt,                          
    bottom = 5pt,                        
]
\fontsize{8.6pt}{13pt}\selectfont
    $<$\textit{\textbf{Atomic}}$>$$::= q \in Q_{atomic}$ $|$ $($$<$\textit{\textbf{List}}$>$$)$ \\
    $<$\textit{\textbf{Dependent}}$>$$::=$$<$\textit{\textbf{Atomic}}$>$$|$$<$\textit{\textbf{Dependent}}$>$$\times$$<$\textit{\textbf{Atomic}}$>$ \\
    $<$\textit{\textbf{List}}$>$$::=$$<$\textit{\textbf{Dependent}}$>$$|$$<$\textit{\textbf{List}}$>$$+$$<$\textit{\textbf{Dependent}}$>$ \\
    $<$\textit{\textbf{Complex}}$>$$::=$ $<$\textit{\textbf{List}}$>$
\end{tcolorbox}

Each rule is necessary to express the target query language:

1. Removing \textbf{$<$\textit{Dependent Query}$>$} makes it impossible to generate queries connected by `$\times$'.

2. Removing \textbf{$<$\textit{List Query}$>$} makes it impossible to generate queries connected by `$+$'.

3. Removing \textbf{$<$\textit{Complex Query}$>$} removes the start symbol of the production rule.

Therefore, no rule is redundant or replaceable.

\subsection*{Nonequivalence of simpler syntax}
We attempt to construct a simpler syntax:

1. If \textbf{$<$\textit{Dependent Query}$>$} is replaced directly with \textbf{$<$\textit{Atomic Query}$>$}, queries with dependencies involving `$\times$' cannot be expressed.

2. If \textbf{$<$\textit{List Query}$>$} is omitted, queries without dependencies involving `$+$' cannot be expressed.
These simplifications fail to generate the target query language, confirming that there is no simpler equivalent grammar.

\subsection{Conclusion}
The grammar is minimal, as it generates the target query language, contains no redundant symbols or rules, and cannot be simplified without reducing its ability of producing complex queries.

\section{Comparison between QCompiler and QDMR}
Based on the results and evaluation of existing benchmarks, our method closely resembles QDMR~\cite{wolfson2020break} in terms of performance representation, because both methods use a seq2seq model to learn decomposition rules to break down multiple search intents of complex questions. However, through deeper syntactical analysis, we observed that many of the operators defined in QDMR, such as \textbf{PROJECT}, \textbf{FILTER}, \textbf{AGGREGATE}, \textbf{BOOLEAN}, and \textbf{COMPARATIVE}, while fully capable of capturing all search intents of the original query, are actually forming a redundant grammar $G' \subseteq G$. From a linguistic standpoint, these operators form a complete but not a minimal grammar, which may be poor for optimization. Our minimal BNF grammar $G[q] \subset G'$, it only defines dependent and parallel relationships via `$\times$' and `$+$', making it clear and easy to represent original complex queries.

Moreover, QDMR specifically trains a model to generate the final answer, which contrasts with the design philosophy of QCompiler. Our aim is to provide a plug-and-play query parsing module within the RAG system, rather than adapting the existing answer format from the benchmark. (6) This approach fosters greater flexibility and allows the components to be reused by the open source community and in practical development.

\section{Algorithm of Query AST Validation}
\label{algo:valid}
\begin{algorithm}[H]
\caption{Query AST Validation Algorithm}
\begin{algorithmic}[1]
\REQUIRE A Root Node node
\ENSURE Returns \textbf{True} if the query is valid, \textbf{False} otherwise
\IF{\text{node.type} = \text{`$<$\textit{\textbf{Atomic Query}}$>$'}}
    \IF{flag = 0 \textbf{and} node.placeholder exists}
        \RETURN \textbf{False}
    \ENDIF
    \IF{flag = 1 \textbf{and not} node.placeholder exists}
        \RETURN \textbf{False}
    \ENDIF
    \RETURN \textbf{True}
\ENDIF
\IF{node.type = \text{`$<$\textit{\textbf{Dependent Query}}$>$'}}
    \STATE Let \( \text{left} \gets \text{node.children[0]} \)
    \STATE Let \( \text{right} \gets \text{node.children[1]} \)
    \RETURN valid\_query(left, flag = 0) and valid\_query(right, flag = 1)
\ENDIF
\IF{node.type = \text{`$<$\textit{\textbf{List Query}}$>$'}}
    \FOR{\( \text{child} \in \text{node.children} \)}
        \IF{\textbf{not} valid\_query(child, flag = flag)}
            \RETURN \textbf{False}
        \ENDIF
    \ENDFOR
    \RETURN \textbf{True}
\ENDIF
\end{algorithmic}
\end{algorithm}

\section{Statistical Description of Four Benchmarks}
\label{appendix:stat}
We provide a statistical description of the four evaluation benchmarks in Table \ref{tab:multi_hop_qa}. 
\begin{table}[h]
    \centering
    \scalebox{.82}{
    \begin{tabular}{l|c|c|c|c}
    \toprule
    \textbf{Name} & \textbf{Source} & \textbf{Train} & \textbf{Valid} & \textbf{Test} \\ \midrule
    HotpotQA & wiki & 90,447 & 7,405 & / \\ 
    2WikiMultiHopQA & wiki & 15,000 & 12,576 & / \\ 
    Musique & wiki & 19,938 & 2,417 & / \\ 
    Bamboogle & wiki & / & / & 125 \\ \bottomrule
    \end{tabular}
    }
    \caption{Multi-hop QA Datasets}
    \label{tab:multi_hop_qa}
\end{table}

For each of HotpotQA, 2WikiMultiHopQA, and Musique, we select the first 1,000 samples from the valid set for evaluation, while for Bamboogle, we use the entire test set for evaluation.

\section{Training Details of Query Compiler}
\label{appendix:train}
We build an instruction-tuning dataset from HotpotQA, 2WikiMultiHopQA, and Musique to train the \textit{Query Expression Translator}. Using Chain-of-Thought prompting and few-shot examples (which is presented in Table \ref{tab:generate_data_prompt}), we prompt the Qwen2.5-72B-Instruct model to generate valid expressions for each query in the training set. 

Next, we use the algorithm described in Appendix \ref{algo:valid} to validate the syntax trees parsed from the expressions, filtering out invalid expressions and get valid query-expression pairs for training.

We fine-tune the base model for $1$ epoch using LoRA~\cite{hu2021lora}, setting the $r=16$, $\alpha = 32$, batch\_size $= 64$, and the learning rate to $5e-5$.

\section{Experiment Implementations}
\label{appendix:implementation}

\textbf{Corpus and Retriever} We use Wikipedia dump 2018~\cite{karpukhin2020dense} as retrieved corpus and bge-base-en-v1.5~\cite{bge_embedding} as dense retriever. For each retrieval process, we concatenate instruction `\textit{Represent this sentence to search relevant passages:}' and query to retrieve relevant top-k documents.

\noindent\textbf{Generator} We use Qwen2.5-7B-Instruct as the base generator to give a response to every query.

\noindent\textbf{Metric} The three evaluation metrics we use are as follows:

(1) Exact Match (EM): Measures whether the generated answer exactly matches the golden answer, ensuring strict consistency.

(2) Accuracy (Acc): Evaluates whether the golden answer is included within the generated answer.

(3) F1 Score: Computes the F1 score between the tokenized generated answer and the tokenized golden answer.

\noindent\textbf{Baseline Settings} 

\textbf{Self-RAG}~\cite{selfrag} We set max\_depth = 2 and beam\_width = 2, threshold = 0.2 for testing. Due to the output format of this method, the EM metric may be underestimated.

\textbf{IR-CoT \& Iter-RetGen}~\cite{ircot,iteretgen} We set max\_iterations = 5 for testing. We found that increasing the number of iterations for these methods does not lead to better performance, so we set a maximum number of iterations accordingly.

\textbf{RQ-RAG}~\cite{rqrag} We set max\_depth = 4 for 2WikiMultihopQA and Musique, and $3$ for HotpotQA and Bamboogle for testing. Since the number of sampled paths grows exponentially with depth and the token count for context increases significantly, we did not experiment with larger maximum depth settings. 

\section{Efficiency Analysis}
In this section, we analysis efficiency of QCompiler from two perspectives: token consumption and throughput.

\textbf{Token Consumption} The efficiency of RAG systems depends on various factors such as CPU, GPU, and storage of the machine. To highlight the efficiency of QCompiler, we focus on token consumption. Each method’s token usage can be divided into three components: prompt tokens, retrieved document tokens, and response generation tokens.

QCompiler improves efficiency by pre-compiling an Abstract Syntax Tree for each query, making the process deterministic and eliminating redundant iterations. This avoids unnecessary prompt and response generation tokens that are common in LLM-based RAG systems. Moreover, The atomicity of the sub-queries in the leaf nodes ensures precise document retrieval and answer generation while maintaining competitive performance. This feature actually improves the efficiency.

We compare the average token consumption per query across different methods when benchmark performance is similar:

\begin{table}[h]
    \centering
    \scalebox{.68}{
    \begin{tabular}{lcccc}
        \toprule
        Methods & 2wikimultihopqa & hotpotqa & musique & bamboogle \\
        \midrule
        Self-RAG & 2079.1 & 2034.2 & 1980.0 & 1945.8 \\
        Iter-Retgen & 3628.0 & 2835.8 & 3430.1 & 3369.4 \\
        IR-CoT & 2462.0 & 1917.9 & 2347.3 & 2320.4 \\
        \rowcolor[RGB]{236,244,252} \textbf{QCompiler} & \textbf{1176.3} & \textbf{1042.0} & \textbf{1061.7} & \textbf{1040.6} \\
        \bottomrule
    \end{tabular}
    }
    \caption{Token consumption of different methods.}
    \label{tab:methods_comparison}
\end{table}

\textbf{Throughput} The baseline methods use LLMs to generate new query content in each iteration, which limits their ability to execute independent sub-queries in parallel. For example, the query “Who is older, James Cameron or Steven Allan Spielberg?” includes two independent sub-queries, but existing methods must process them sequentially. In contrast, our compiled AST allows for parallel execution of child nodes in list queries, significantly improving throughput. This feature also improves efficiency.

\section{Open-domain Question with QCompiler}
Open-domain question answering is a current challenge in research, and the quality of the answers largely depends on the performance of the base model. For example, the implicit comparative question (e.g. Compare the market share and revenue growth of top 5 EV manufacturers in North America and Europe over the last 3 years.) may need agentic RAG systems to autonomously expand search intents during running time, and the current baselines have not been able to fully address these issues. To solve this problem, QCompiler can be integrated into agentic RAG systems to generate a AST for the remaining question in each iteration, but this is beyond the scope of this work.

In this context, our method still attempts to generate several sub-intents of the original query, retrieving relevant documents and integrating information to provide a reference for the final answer, which is shown in Table \ref{tab:example_q1} and Table \ref{tab:example_q2}.

\section{Other Disccusions}
We found that, under the grammar instructions $G[q]$ and the few-shot examples format, the 7B model already demonstrated relatively strong parsing capabilities. Although our experiments were conducted by building training data from the training sets provided by three benchmarks to fine-tune smaller models, this step was actually aimed at obtaining a specialized model with fewer parameters for the challenge of low-resource settings.

Similarly, the focus of this work is to design a minimal, and sufficient structured intermediate query representation for RAG systems. Therefore, we did not train the base model for generating the final answers, and hence avoided overfitting to the existing benchmarks.

\begin{table*}[!t]
    \centering
    \caption{Prompt to generate training data}
    \fontsize{9pt}{11pt}\selectfont
    \begin{tabular}{p{0.98\linewidth}}
    \midrule
        \rowcolor{gray!20}\textbf{Chain-of-Thought Prompting} \\
    \midrule
\textbf{You are an expert in query intent understanding, tasked with decomposing complex queries into basic components. Follow the step-by-step procedure below to generate a BNF-compliant expression for each query.} \\

========================================== \\
Query Types and Grammar Definitions \\
========================================== \\
Query Types: \\
    \ \ \ \ 1. AtomicQuery: \\
       \ \ \ \ \ \ \ \ • Simple, direct queries that require a factual answer. \\
       \ \ \ \ \ \ \ \ • Non-decomposable, orthogonal, and non-redu
    \ \ \ \ 2. DependentQuery: \\
       \ \ \ \ \ \ \ \ • Multi-step queries where each step depends on the result of the previous one. \\
       \ \ \ \ \ \ \ \ • Composed of multiple AtomicQueries with dependencies. \\
    \ \ \ \ 3. ListQuery: \\
       \ \ \ \ \ \ \ \ • Requires decomposition into multiple parallel, independent sub-queries. \\
BNF Definitions: \\
    \ \ \ \ • Set of atomic query terms (W): \\
         \ \ \ \ \ \ \ \ All possible atomic query strings, where each atomic query is independent and non-redundant. \\
    \ \ \ \ • Set of operators (O): {'+' (parallel), '*' (dependent)} \\
    \ \ \ \ • <AtomicQuery> ::= w $\in$ W | '(' <ListQuery> ')' \\
        \ \ \ \ \ \ \ \ - expressions enclosed in parentheses can also be considered as a production rule for <AtomicQuery>. \\
    \ \ \ \ • <DependentQuery> ::= <AtomicQuery> | <DependentQuery> '*' <AtomicQuery> \\
        \ \ \ \ \ \ \ \ - <AtomicQuery> may include placeholders formatted as {placeholder name}. \\
        \ \ \ \ \ \ \ \ - '*' indicates that the next query depends on the result of the previous query. \\
    \ \ \ \ • <ListQuery> ::= <DependentQuery> | <ListQuery> '+' <DependentQuery> \\
        \ \ \ \ \ \ \ \ - '+' denotes parallel relationships among queries. \\
========================================== \\
 Example Workflows \\
========================================== \\
--- Example --- \\
query = How many Germans live in the colonial holding in Aruba's continent that was governed by Prazeres's country? \\

Step1: **Define atomic queries:** \\
  \ \ \ \ - query1.1: Which continent is Aruba in? \\
  \ \ \ \ - query1.2: Which country is Prazeres in? \\
  \ \ \ \ - query2: Which colonial holding in \{continent\} was governed by \{country\}? \\
  \ \ \ \ - query3: How many Germans live in \{colonial\_holding\}? \\

Step2: **Queries Combination:** \\
  \ \ \ \ Thought: \\
    \ \ \ \ \ \ \ \ • query1.1 and query1.2 are parallel → Use '+' \\
    \ \ \ \ \ \ \ \ • query2 depends on the results of both query1.1 and query1.2 → Use '*' \\
    \ \ \ \ \ \ \ \ • query3 depends on query2 → Use '*' \\
  \ \ \ \ - Combine: \\
    \ \ \ \ \ \ \ \ (query1.1 + query1.2) * query2 * query3 \\

compiled\_expression = (Which continent is Aruba in? + Which country is Prazeres in?) * Which colonial holding in \{continent\} was governed by \{country\}? * How many Germans live in \{colonial\_holding\}? \\

========================================== \\
 Task Instructions \\
========================================== \\
\ \ \ \ 1. Decompose the user's query into an ordered set of AtomicQueries, ensuring each query is factual, independent, and non-redundant. \\
\ \ \ \ 2. Determine whether sub-queries are parallel (use '+') or dependent (use '*'). \\
\ \ \ \ 3. If you use '*', the next query must have a placeholder referencing the previous step’s result, e.g., \{placeholder\}. \\
\ \ \ \ 4. Output your reasoning in two steps: \\
   \ \ \ \ \ \ \ \ • Step1: **Define atomic queries** \\
   \ \ \ \ \ \ \ \ • Step2: **Queries Combination** \\
   \ \ \ \ \ \ \ \ Then provide the final expression. \\
\ \ \ \ 5. Maintain the same language as the input query when formulating AtomicQueries. \\
\ \ \ \ 6. Follow the example format to ensure consistency. \\
Please decompose and compile each complex query into a BNF-compliant expression using '+', '*', '()', and '\{placeholders\}', then output in the specified format. \\
    \bottomrule
    \end{tabular}
    \label{tab:generate_data_prompt}
\end{table*}

\begin{table*}[!t]
    \centering
    \caption{An example of QComplier's Query Expression Translator and parsed AST}
    \fontsize{9pt}{11pt}\selectfont
    \begin{tabular}{p{0.98\linewidth}}
    \midrule
        \rowcolor{gray!20}\textbf{System Prompt} \\
    \midrule
        \textbf{You are an expert in query intent understanding, tasked with decomposing complex queries into basic components. Follow the step-by-step procedure below to generate a BNF-compliant expression for each query.} \\
        =================================== \\
 Query Types and Grammar Definitions \\
=================================== \\

Query Types: \\
    \ \ \ \ 1. AtomicQuery: \\
       \ \ \ \ \ \ \ \ • Simple, direct queries that require a factual answer. \\
       \ \ \ \ \ \ \ \ • Non-decomposable, orthogonal, and non-redundant. \\

    \ \ \ \ 2. DependentQuery: \\
       \ \ \ \ \ \ \ \ • Multi-step queries where each step depends on the result of the previous one. \\
       \ \ \ \ \ \ \ \ • Composed of multiple AtomicQueries with dependencies. \\

    \ \ \ \ 3. ListQuery: \\
       \ \ \ \ \ \ \ \ • Requires decomposition into multiple parallel, independent sub-queries. \\

BNF Definitions: \\
    \ \ \ \ • Set of atomic query terms (W): \\
         \ \ \ \ \ \ \ \ All possible atomic query strings, where each atomic query is independent and non-redundant. \\

    \ \ \ \ • Set of operators (O): {`+' (parallel), `*' (dependent)} \\

    \ \ \ \ • <AtomicQuery> ::= w $\in$ W | `(' <ListQuery> `)' \\
        \ \ \ \ \ \ \ \ - expressions enclosed in parentheses can also be considered as a production rule for <AtomicQuery>. \\

    \ \ \ \ • <DependentQuery> ::= <AtomicQuery> | <DependentQuery> `*' <AtomicQuery> \\
        \ \ \ \ \ \ \ \ - <AtomicQuery> may include placeholders formatted as {placeholder name}. \\
        \ \ \ \ \ \ \ \ - `*' indicates that the next query depends on the result of the previous query. \\

    \ \ \ \ • <ListQuery> ::= <DependentQuery> | <ListQuery> `+' <DependentQuery> \\
        \ \ \ \ \ \ \ \ - `+' denotes parallel relationships among queries. \\
        
==================================== \\
 Task Instructions \\
==================================== \\
\ \ \ \ 1. Decompose the user's query into an ordered set of AtomicQueries, ensuring each query is factual, independent, and non-redundant. \\
\ \ \ \ 2. Determine whether sub-queries are parallel (use `+') or dependent (use `*'). \\
\ \ \ \ 3. If you use `*', the next query must have a placeholder referencing the previous step’s result, e.g., \{placeholder\}. \\
\ \ \ \ 4. Maintain the same language as the input query when formulating AtomicQueries. \\

Please decompose and compile each complex query into a BNF-compliant expression using `+', `*', `()', and `\{placeholders\}', then output in the specified format. \\
    \midrule
        \rowcolor{gray!20}
        \textbf{Input (Original Query)} \\
    \midrule
        I want to find an introduction and reviews of JK. Rowling’s most popular book and checking if the local library has it? \\
    \midrule
        \rowcolor{gray!20} \textbf{Output (BNF Expression)} \\
    \midrule
        What is JK. Rowling’s most popular book? * (Find an introduction to \{book\} + Find reviews of \{book\} + Does the local library have \{book\}?) \\
    \midrule 
        \rowcolor{gray!20} \textbf{Parsed AST} \\
    \midrule 
\textbf{ComplexQuery}(value=`What is JK. Rowling’s most popular book? * (Find an introduction to \{book\} + Find reviews of \{book\} + Does the local library have \{book\}?)', [ \\
  \ \ \ \ \textbf{DependentQuery}(value=`What is JK. Rowling’s most popular book? * (Find an introduction to \{book\} + Find reviews of \{book\} + Does the local library have \{book\}?)', [ \\ 
    \ \ \ \ \ \ \ \ \textbf{AtomicQuery}(value=`What is JK. Rowling’s most popular book?'), \\
    \ \ \ \ \ \ \ \ \textbf{ListQuery}(value='(Find an introduction to \{book\} + Find reviews of \{book\} + Does the local library have \{book\}?)', placeholder=[`book'], [ \\
      \ \ \ \ \ \ \ \ \ \ \ \ \textbf{AtomicQuery}(value=`Find an introduction to \{book\}', placeholder=[`book']), \\
      \ \ \ \ \ \ \ \ \ \ \ \ \textbf{AtomicQuery}(value=`Find reviews of \{book\}', placeholder=[`book']), \\
      \ \ \ \ \ \ \ \ \ \ \ \ \textbf{AtomicQuery}(value=`Does the local library have \{book\}?', placeholder=[`book']) \\
    \ \ \ \ \ \ \ \ ]) \\
  \ \ \ \ ]) \\
]) \\
    \bottomrule
    
    \end{tabular}
    \label{tab:AST}
\end{table*}

\begin{table*}[!t]
    \centering
    \caption{An example for \textbf{multi-hop} question with QCompiler}
    \fontsize{9pt}{11pt}\selectfont
    \begin{tabular}{p{0.98\linewidth}}
    \midrule
        \rowcolor{gray!20}\textbf{Example \#1 Multi-hop question from Musique} \\
    \midrule
        \textbf{Question:} \\
        Why did Roncalli leave the city where the creator of La Schiavona died? \\
        \textbf{Ground truth:} \\
        for the conclave in Rome; Rome; Roma \\
    \midrule
        \rowcolor{gray!20}
        \textbf{BNF Expression} \\
    \midrule
        Who is the creator of La Schiavona? * Where did \{creator\} die? * Why did Roncalli leave \{city\}? \\
    \midrule
        \rowcolor{gray!20}
        \textbf{Parsed AST} \\
    \midrule
        \textbf{ComplexQuery}(value=`Who is the creator of La Schiavona? * Where did \{creator\} die? * Why did Roncalli leave \{city\}?', [ \\
    \ \ \ \ \textbf{DependentQuery}(value=`Who is the creator of La Schiavona? * Where did \{creator\} die? * Why did Roncalli leave \{city\}?', [ \\
        \ \ \ \ \ \ \ \ \textbf{DependentQuery}(value=`Who is the creator of La Schiavona? * Where did \{creator\} die?', [ \\
            \ \ \ \  \ \ \ \  \ \ \ \  \textbf{AtomicQuery}(value=`Who is the creator of La Schiavona?'), \\
            \ \ \ \ \ \ \ \ \ \ \ \ \textbf{AtomicQuery}(value=`Where did \{creator\} die?', placeholder=[`creator']) \\
        ]), \\
        \ \ \ \ \ \ \ \ \textbf{AtomicQuery}(value=`Why did Roncalli leave \{city\}?', placeholder=[`city']) \\
    \ \ \ \ ]) \\
]) \\
    \midrule
        \rowcolor{gray!20}
        \textbf{Depth-Fisrt-Search Process in AST} \\
    \midrule
        \textbf{AtomicQuery 1:} \\
        Who is the creator of La Schiavona? \\
        \textbf{Retrieved Documents:} \\
        Doc1, Title = Andrea Schiavone Andrea Meldolla, Content = ... \\
        Doc2, Title = ..., Content = ... \\
        ... \\
        \textbf{Result:} \\
        The creator of La Schiavona is Titian. This information can be directly inferred from the document titled \"La Schiavona,\" which states, \"Portrait of a Lady, also known as La Schiavona (the woman from Dalmatia), is a 1510–12 portrait by Titian of an unknown woman. \\ \\
        \textbf{AtomicQuery 2:} \\
        Where did \textbf{Titian} die? (get placeholder `creator''s value from its left node.)\\
        \textbf{Retrieved Documents:} \\
        ... \\
        \textbf{Result:} \\
        According to the documents, Titian died in Venice on 27 August 1576. \\ \\
        \textbf{AtomicQuery 3:} \\
        Why did Roncalli leave \textbf{Venice}? (get placeholder `city''s value from its left node.) \\
        \textbf{Retrieved Documents:} \\
        ... \\
        \textbf{Result:} \\
        Based on the information provided in the documents, Roncalli left Venice after the death of Pope Pius XII on 9 October 1958 to participate in the papal conclave in Rome. Doc2 mentions, \"Following the death of Pope Pius XII on 9 October 1958, Roncalli watched the live funeral on his last full day in Venice on 11 October. His journal was specifically concerned with the funeral and the abused state of the late pontiff's corpse. Roncalli left Venice for the conclave in Rome well aware that he was `papabile', and after eleven ballots, was elected to. \\
    \midrule
        \rowcolor{gray!20}
        \textbf{Concatenated three sub-queries and answers to generate the final answer} \\
    \midrule
    Answer the question based the original question and its sub-questions and answers. \\
    \{orignal question\} \\
    \{sub-qurstions and answers\} \\

    \rowcolor[RGB]{236,244,252}\textbf{Answer}: Because Pope Pius XII died on 9 October 1958, Roncalli left Venice to participate in the papal \textbf{conclave in Rome}. \\
    \bottomrule
    \end{tabular}
    \label{tab:example_multihop}
\end{table*}

\begin{table*}[!t]
    \centering
    \caption{An example for \textbf{summarization} question with QCompiler}
    \fontsize{9pt}{11pt}\selectfont
    \begin{tabular}{p{0.98\linewidth}}
    \midrule
        \rowcolor{gray!20}\textbf{Example \#2 Summarization Question} \\
    \midrule
        \textbf{Question:} \\
        Given the current geopolitical tensions and supply chain disruptions, how might the semiconductor industry evolve over the next five years?\\
    \midrule
        \rowcolor{gray!20}
        \textbf{BNF Expression} \\
    \midrule
        What are the current geopolitical tensions affecting the semiconductor industry + What are the current supply chain disruptions affecting the semiconductor industry + How might the semiconductor industry adapt to the geopolitical tensions over the next five years + How might the semiconductor industry adapt to the supply chain disruptions over the next five years \\
    \midrule
        \rowcolor{gray!20}
        \textbf{Parsed AST} \\
    \midrule
    \textbf{ComplexQuery}(value=`What are the current geopolitical tensions affecting the semiconductor industry + What are the current supply chain disruptions affecting the semiconductor industry + How might the semiconductor industry adapt to the geopolitical tensions over the next five years + How might the semiconductor industry adapt to the supply chain disruptions over the next five years', [ \\
    \ \ \ \ \textbf{AtomicQuery}(value=`What are the current geopolitical tensions affecting the semiconductor industry'), \\
    \ \ \ \ \textbf{AtomicQuery}(value=`What are the current supply chain disruptions affecting the semiconductor industry'), \\
    \ \ \ \ \textbf{AtomicQuery}(value=`How might the semiconductor industry adapt to the geopolitical tensions over the next five years'), \\
    \ \ \ \ \textbf{AtomicQuery}(value=`How might the semiconductor industry adapt to the supply chain disruptions over the next five years') \\
]) \\
    \midrule
        \rowcolor{gray!20}
        \textbf{Concatenated three sub-queries and answers to generate the final answer} \\
    \midrule
        \rowcolor[RGB]{236,244,252} The semiconductor industry might evolve by diversifying supply chains, forming public-private partnerships, engaging in collaborative efforts, increasing investment in R\&D, focusing on advanced technologies, and adapting to market volatility. It will also enhance supply network resilience through dynamic business model reinvention, collaboration across the network, maintaining a flexible supply chain, and developing strategic resilience. \\
    \bottomrule
    \end{tabular}
    \label{tab:example_q1}
\end{table*}

\begin{table*}[!t]
    \centering
    \caption{An example for \textbf{implicit comparative} question with QCompiler}
    \fontsize{9pt}{11pt}\selectfont
    \begin{tabular}{p{0.98\linewidth}}
    \midrule
        \rowcolor{gray!20}\textbf{Example \#3 Implicit Comparative Question} \\
    \midrule
        \textbf{Question:} \\
        Compare the market share and revenue growth of top 5 EV manufacturers in North America and Europe over the last 3 years. \\
    \midrule
        \rowcolor{gray!20}
        \textbf{BNF Expression} \\
    \midrule
        Who are the top 5 EV manufacturers in North America * What is the market share and revenue growth of \{manufacturer\} in North America over the last 3 years + Who are the top 5 EV manufacturers in Europe * What is the market share and revenue growth of \{manufacturer\} in Europe over the last 3 years \\
    \midrule
        \rowcolor{gray!20}
        \textbf{Parsed AST} \\
    \midrule
    \textbf{ComplexQuery}(value=`Who are the top 5 EV manufacturers in North America * What is the market share and revenue growth of \{manufacturer\} in North America over the last 3 years + Who are the top 5 EV manufacturers in Europe * What is the market share and revenue growth of \{manufacturer\} in Europe over the last 3 years', [ \\
    \ \ \ \ \textbf{DependentQuery}(value=`Who are the top 5 EV manufacturers in North America * What is the market share and revenue growth of \{manufacturer\} in North America over the last 3 years', [ \\
        \ \ \ \ \ \ \ \ \textbf{AtomicQuery}(value=`Who are the top 5 EV manufacturers in North America'), \\
        \ \ \ \ \ \ \ \ \textbf{AtomicQuery}(value=`What is the market share and revenue growth of \{manufacturer\} in North America over the last 3 years', placeholder=[`manufacturer']) \\
    \ \ \ \ ]), \\
    \ \ \ \ \textbf{DependentQuery}(value=`Who are the top 5 EV manufacturers in Europe * What is the market share and revenue growth of \{manufacturer\} in Europe over the last 3 years', [ \\
        \ \ \ \ \ \ \ \ \textbf{AtomicQuery}(value=`Who are the top 5 EV manufacturers in Europe'), \\
        \ \ \ \ \ \ \ \ \textbf{AtomicQuery}(value=`What is the market share and revenue growth of \{manufacturer\} in Europe over the last 3 years', placeholder=[`manufacturer']) \\
    \ \ \ \ ]) \\
]) \\
    \bottomrule
    \end{tabular}
    \label{tab:example_q2}
\end{table*}

\end{document}